%% file: iclr2027_conference.tex
\documentclass{article}

\usepackage{iclr2027_conference,times}

\input{math_commands.tex}

\usepackage{amssymb}

\usepackage{hyperref}
\usepackage{url}
\usepackage{graphicx}
\usepackage{wrapfig}
\usepackage{booktabs}
\usepackage{multirow}
\usepackage[dvipsnames]{xcolor}

\makeatletter
\providecommand{\wrapfill}{%
  \par
  \ifnum\c@WF@wrappedlines>1
    \leavevmode
    \vspace{\dimexpr\c@WF@wrappedlines\baselineskip\relax}\par
  \fi
}
\makeatother

\usepackage{letterspace}
\usepackage{etoc}

\newcommand{\appendixopening}{%
  \clearpage
  \begingroup
  \centering
  \noindent\rule{\textwidth}{1.1pt}\par
  \vspace{1.1em}
  {\normalsize\scshape
    \textls[180]{Supplementary Material}\par}
  \vspace{0.7em}
  {\Huge\bfseries\textls[120]{Appendix}\par}
  \vspace{1.1em}
  \noindent\rule{\textwidth}{0.5pt}\par
  \endgroup
  \vspace{2.2em}
}

\etocsettocstyle
  {{\centering\large\bfseries Contents\par}\vspace{1.2em}}
  {}

\definecolor{darkgreen}{RGB}{0,128,0}

\definecolor{peizhiorange}{RGB}{214,107,17}

\definecolor{unlearnblue}{RGB}{220,239,250}
\definecolor{utilityorange}{RGB}{255,231,216}
\definecolor{uesgreen}{RGB}{0,128,64}
\definecolor{uesred}{RGB}{190,35,45}

\newcommand{\metrichead}[2]{%
  \begingroup
  \setlength{\fboxsep}{1.5pt}%
  \colorbox{#1}{\strut #2}%
  \endgroup
}

\newcommand{\ueschange}[4]{%
  #1\,{\scriptsize\textcolor{#4}{(#2\%$#3$)}}%
}

\title{Making LLMs Truly Forget: Deep Unlearning by Searching, Selecting, and Severing Knowledge Paths}

\author{%
  \textbf{Jialu Wang}$^{1}$ \quad
  \textbf{Peizhi Niu}$^{2}$ \quad
  \textbf{Haoteng Yin}$^{3}$ \quad
  \textbf{Hans Hao-Hsun Hsu}$^{4}$\quad
  \textbf{Pan Li}$^{4}$ \quad
  \textbf{Rongzhe Wei}$^{4}$\thanks{%
    Corresponding author:
    \href{mailto:rongzhe.wei@gatech.edu}
         {\texttt{rongzhe.wei@gatech.edu}}.%
  } \\[0.8em]
  $^{1}$Tongji University \quad
  $^{2}$University of Illinois Urbana-Champaign \quad
  $^{3}$Purdue University \\
  $^{4}$Georgia Institute of Technology
}
\iclrfinalcopy

\begin{document}

\maketitle

\pagestyle{plain}
\thispagestyle{plain}

\etocdepthtag.toc{mtchapter}

\input{sections/0-abstract}

\input{sections/1-introduction}

\input{sections/2-related-work}
\input{sections/3-formulation}
\input{sections/4-methodology}

\input{sections/5-evaluation}
\input{sections/6-experiments}

\input{sections/7-results}
\input{sections/8-conclusion}

\clearpage
\input{sections/9-ai-use-statement}
\clearpage

\bibliographystyle{iclr2027_conference}
\bibliography{iclr2027_conference}

\appendix

\appendixopening

\etocdepthtag.toc{mtappendix}
\etocsettagdepth{mtchapter}{none}
\etocsettagdepth{mtappendix}{subsection}
\tableofcontents
\newpage

\input{sections/A-appendix-experimental-settings}
\input{sections/B-appendix-dataset-construction}
\input{sections/C-appendix-kg-statistics}
\input{sections/D-appendix-search-efficiency}
\input{sections/E-appendix-entropy-mapping}
\input{sections/F-appendix-additional-ablations}

\input{sections/99-floating-notes}
\end{document}

%% file: math_commands.tex
\usepackage{amsmath,amsfonts,bm}

\def\eqref#1{equation~\ref{#1}}

\def\1{\bm{1}}

\DeclareMathAlphabet{\mathsfit}{\encodingdefault}{\sfdefault}{m}{sl}
\SetMathAlphabet{\mathsfit}{bold}{\encodingdefault}{\sfdefault}{bx}{n}



%% file: sections/0-abstract.tex

\begin{abstract}

While an unlearned language model may no longer recall a fact directly, the fact often remains recoverable through multi-hop reasoning over related knowledge. Most existing unlearning techniques overlook this vulnerability, targeting facts in isolation while leaving their supporting knowledge intact. To achieve true forgetting, we propose a general deep unlearning framework compatible with existing unlearning algorithms. Our approach adaptively explores both explicit responses and latent internal representations to discover valid reasoning paths, compiles them into a confidence-aware supporting subgraph, and we apply a graph minimum cut to sever all recovery paths while preserving unrelated knowledge. To rigorously evaluate deep unlearning, we introduce a model-specific pipeline that extracts and completes knowledge graphs from raw text, filtering them by calibrated model confidence to reflect what the model genuinely retains. Comprehensive experiments demonstrate that selectively unlearning supporting knowledge yields substantially deeper forgetting than superficial methods while preserving model utility, highlighting that genuine unlearning requires breaking the relational structures that enable factual reconstruction.

\end{abstract}

%% file: sections/1-introduction.tex
\section{Introduction}

Large language models (LLMs) acquire broad knowledge and capabilities from large-scale text corpora. However, these corpora may also contain copyrighted material or private information that models should not retain~\citep{henderson2023foundation,min2024silo}. Removing the influence of such information after training is challenging, as retraining the model from scratch without the target data is often prohibitively expensive~\citep{eldan2023harry,jang2023knowledge}. Machine unlearning addresses this challenge by efficiently modifying a trained model to remove the influence of designated data while preserving its behavior on unrelated data~\citep{cao2015towards,sekhari2021remember,wei2024underestimated}.

\begin{figure}[t]
    \centering
    \includegraphics[width=\linewidth]{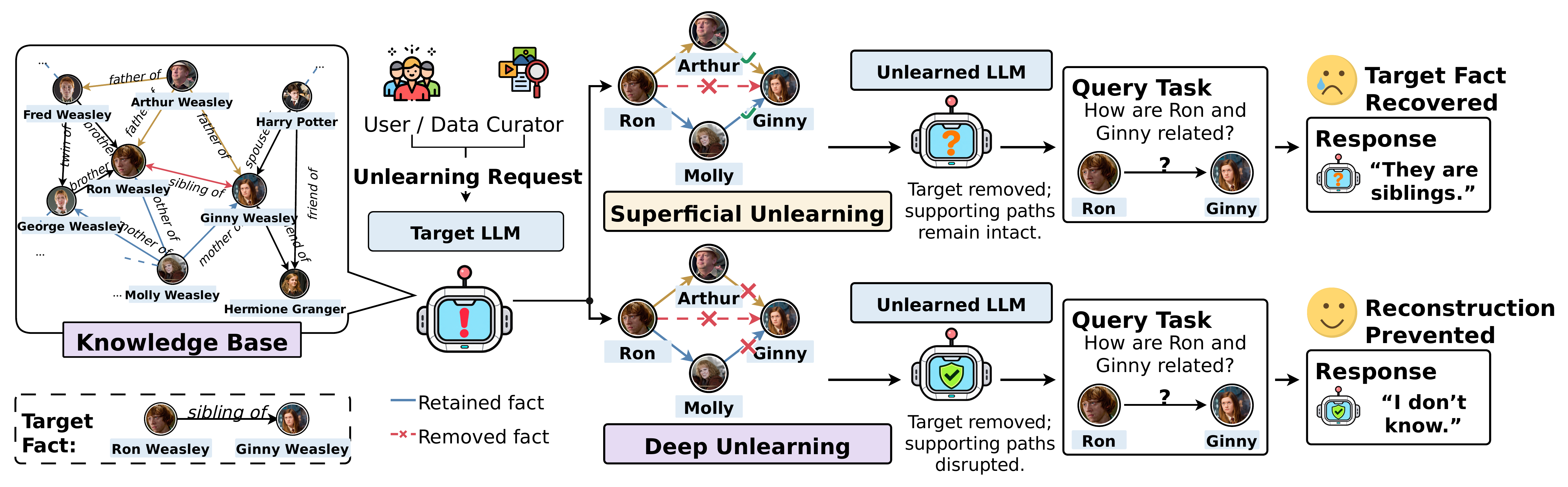}
    \caption{Illustration of deep unlearning. A target fact may remain recoverable through reasoning over related knowledge even after unlearning. Deep unlearning aims to interrupt the supporting knowledge structures that enable such recovery.}
    \label{fig:deepunlearn_intro}
\end{figure}

Despite recent progress in factual unlearning, most existing methods still treat knowledge as a collection of isolated targets~\citep{yao2024large,maini2024tofu,shi2025muse,jin2024rwku,ma2025entity}. However, factual knowledge in LLMs is highly interconnected, and removing direct access to a fact does not necessarily make it unrecoverable: related knowledge may still support its reconstruction~\citep{wei2025forget,wang2026erasing}. For instance, as illustrated in Figure~\ref{fig:deepunlearn_intro}, even if the model no longer directly recalls that ``Ron and Ginny Weasley'' are siblings, it may still retain that both are children of ``Arthur and Molly Weasley'', from which the sibling relation can be inferred. Thus, effective unlearning requires more than removing the target fact itself. It must also account for the supporting knowledge that keeps the target inferable. We refer to this stronger objective as deep unlearning~\citep{wu2024evaluating,wei2025forget}.

To address this limitation, we propose Deep Unlearning via Minimum Inferential Cuts (\textsc{DUMIC}), a general framework compatible with existing unlearning algorithms. At a high level, given a target fact, our framework dynamically maps out both explicit dependencies and implicit associations retained by the model to identify the supporting knowledge that enables reconstruction. Specifically, \textsc{DUMIC} operates in three stages: search, select, and sever. During search, to capture inference paths beyond what surface text reveals, we integrate explicit model responses with latent entities decoded from intermediate-layer representations through a decoding procedure that maps these representations to potentially relevant entities that are excluded from the model’s response. These additional entities expand the search frontier and uncover supporting paths that would otherwise remain unexplored. Using only explicitly mentioned entities may therefore miss paths through which the target fact can be reconstructed. During the select stage, we verify whether these candidate paths genuinely substantiate the target fact, filtering out invalid associations and calibrating model confidence to isolate the active supporting subgraph. Finally, in the severing stage, we formulate supporting-fact selection as a confidence-weighted graph minimum cut over this subgraph. This identifies the minimal-cost subset of facts necessary to break all verified recovery paths, achieving thorough unlearning while strictly preserving model utility.

To rigorously evaluate deep unlearning, we introduce a model-specific evaluation pipeline that assesses whether target knowledge remains recoverable from the facts retained by each model. Grounded in a comprehensive reference graph extracted and completed from raw text, our pipeline filters relational facts through calibrated confidence to mirror each model's actual knowledge state. Extensive experiments reveal three key findings. First, surface model responses conceal critical reasoning routes: latent entities decoded from intermediate representations substantially expand path coverage, exposing hidden avenues for factual reconstruction. Second, superficial unlearning offers only an illusion of forgetting, as leaving supporting knowledge intact allows target facts to remain reliably inferable via multi-hop reasoning. Third, \textsc{DUMIC}'s confidence-aware minimum cut surgically severs these inferential paths, achieving deep forgetting while mitigating the catastrophic utility collapse incurred by unconstrained knowledge removal.

%% file: sections/2-related-work.tex
\section{Related Work}
\paragraph{Superficial Unlearning.}
Most LLM unlearning methods reduce a model's reliance on a designated forget set~\citep{eldan2023harry,kurmanji2023unbounded,liu2024rethinking,jia2024soul,niu2025guard,wang2026quantizationrobust}. Gradient Ascent (GA) suppresses target responses by increasing forget-set loss~\citep{jang2023knowledge,yao2024machine} but risks damaging retained knowledge~\citep{kurmanji2023unbounded,maini2024tofu,jia2024soul}, whereas Negative Preference Optimization (NPO) improves stability by reducing forget-data likelihood relative to the original model~\citep{zhang2024negative}. Benchmarks such as TOFU and MUSE evaluate target accessibility and retained utility through fictitious biographies and memorized text~\citep{eldan2023harry,maini2024tofu,shi2025muse}, but overlook whether related knowledge can reconstruct the target~\citep{wei2025forget,wang2026erasing}.

\paragraph{Deep Unlearning.}
Deep unlearning requires a target fact to become uninferable from retained knowledge, beyond suppressing direct recall~\citep{wu2024evaluating,wei2025forget}. Prior studies use rule-based deduction on fixed knowledge bases~\citep{wu2024evaluating} or subgraphs guided by external graphs~\citep{wei2025forget}, showing that erased facts can persist through logical associations and paraphrases~\citep{wang2026erasing,wu2025learning}. A related line of research studies relational knowledge through \emph{multi-hop knowledge editing}~\citep{zhong2023mquake,shi2024retrieval,cohen2024ripple,cheng2025compke}, but the two paradigms operate in opposite directions. Multi-hop editing follows forward (``cause-to-effect'') propagation, updating an upstream premise and checking its downstream consequences; deep unlearning instead poses a reverse (``effect-to-cause'') challenge, tracing a target fact backward to identify \emph{which} supporting premises must be selectively severed to prevent its reconstruction.

%% file: sections/3-formulation.tex
\section{The Formulation of Knowledge Unlearning}
\label{sec:formulation}

\noindent\textbf{Confidence-Aware Knowledge Modeling in LLMs.}
We represent the relational knowledge of a pretrained LLM $\mathcal{M}_{\mathrm{pre}}$ as a confidence-aware graph $\mathcal{G}_{\mathcal{M}_{\mathrm{pre}}}=(\mathcal{E},\mathcal{R},\mathcal{T}_{\mathcal{U}})$, where $\mathcal{E}$ and $\mathcal{R}$ denote the entity and relation sets, respectively. Each fact $t=(s,r,o,u)\in\mathcal{T}_{\mathcal{U}}$ consists of a subject, a relation, an object, and a calibrated confidence $u$, which reflects the model's confidence in the corresponding triple. Given a target triple $e=(s^{*},r^{*},o^{*})$, the model may retain $e$ directly or reconstruct it from related facts. A supporting path $P_i=(t_{i,1},\ldots,t_{i,\ell_i})$ consists of facts forming a connected reasoning chain that supports $e$. We denote the set of supporting paths retained by the model for target $e$ as $\mathcal{P}_{\mathcal{M}}(e)$. For example, $(\text{Ron},\text{\textsc{SiblingOf}},\text{Ginny})$ may be inferred from $(\text{Arthur},\text{\textsc{FatherOf}},\text{Ron},u_1)$ and $(\text{Arthur},\text{\textsc{FatherOf}},\text{Ginny},u_2)$.

\noindent\textbf{LLM Unlearning.}
Let $\mathcal{M}_{\mathcal{A}}$ be the model obtained by applying an unlearning method $\mathcal{A}$ to $\mathcal{M}_{\mathrm{pre}}$, and let $u_{\mathcal{A}}(t)$ denote its calibrated confidence in a fact $t$. For a target fact $e=(s^{*},r^{*},o^{*})$, let $\mathcal{P}_{\mathcal{M}_{\mathrm{pre}}}(e)$ be its verified supporting paths in the pretrained model, excluding $e$ itself. Because reconstructing $e$ through a path $P$ requires every constituent fact, we define the path's confidence after unlearning as $C_{\mathcal{A}}(P)=\min_{t\in P}u_{\mathcal{A}}(t)$. Let $\mathcal{Q}_e\subseteq\mathcal{P}_{\mathcal{M}_{\mathrm{pre}}}(e)$ denote the supporting paths considered in the unlearning objective. Given a threshold $\gamma$, $\mathcal{A}$ achieves deep unlearning of $e$ with respect to $\mathcal{Q}_e$ if $u_{\mathcal{A}}(e)\leq\gamma$ and $C_{\mathcal{A}}(P)\leq\gamma$ for every $P\in\mathcal{Q}_e$. The complete objective takes $\mathcal{Q}_e=\mathcal{P}_{\mathcal{M}_{\mathrm{pre}}}(e)$, requiring every verified supporting path to be weakened. Superficial unlearning is the special case $\mathcal{Q}_e=\varnothing$, in which the same criterion reduces to $u_{\mathcal{A}}(e)\leq\gamma$. At the subgraph level, a set of supporting facts $C$ cuts all paths under consideration if $C\cap P\neq\varnothing$ for every $P\in\mathcal{Q}_e$.





%% file: sections/4-methodology.tex

\section{Methodology}
\label{sec:methodology}


\subsection{Overview of the Deep Unlearning Framework}
\begin{figure}[t]
    \centering   
    \vspace{-2mm}
    \includegraphics[width=0.95\textwidth]{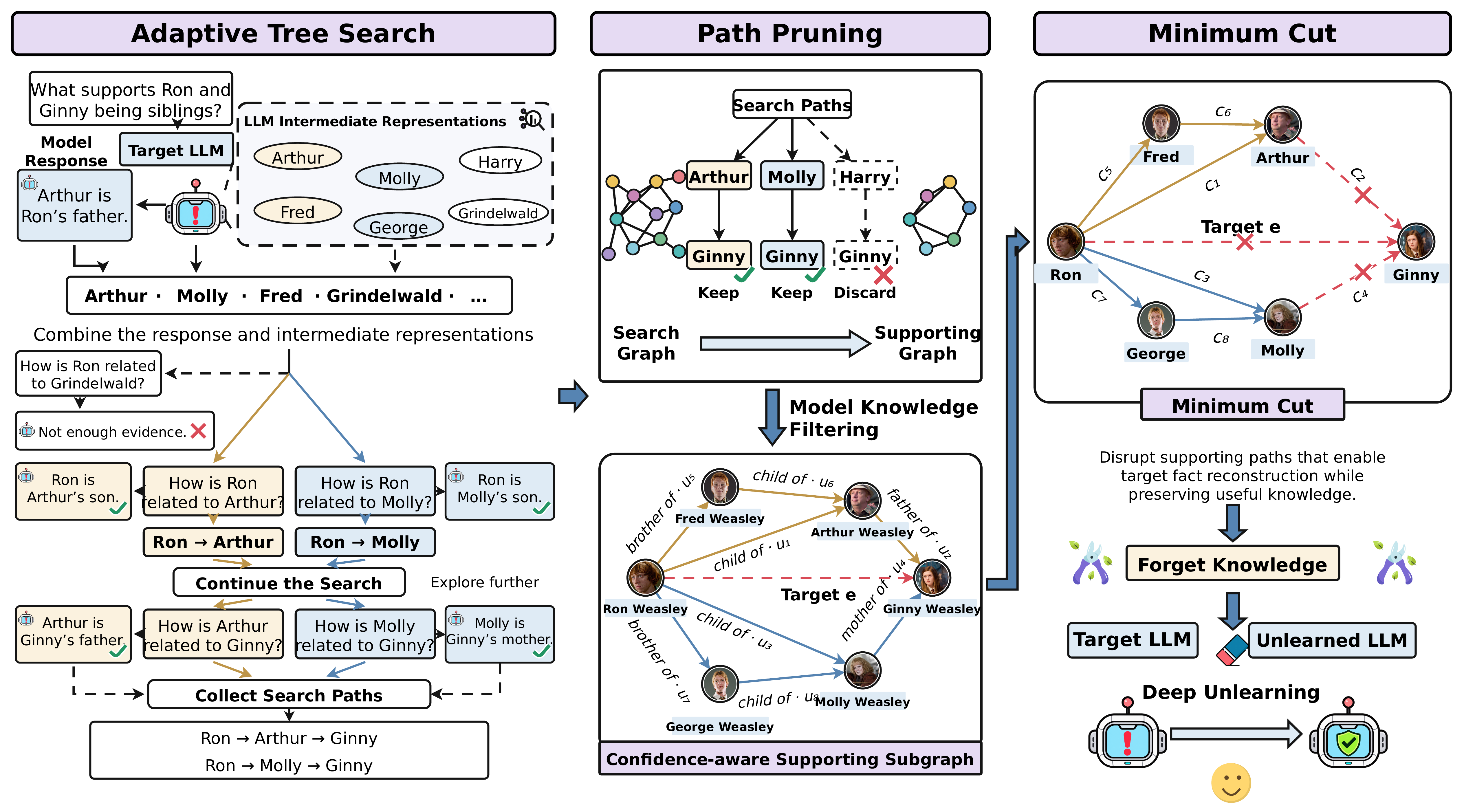}
    \vspace{-4mm}
\caption{Overview of the proposed deep unlearning framework. Adaptive tree search discovers candidate supporting paths, path pruning constructs a confidence-aware supporting subgraph, and minimum-cut selection disrupts these paths while preserving model utility.}
    \label{fig:overview}
\end{figure}

Building on the definition of deep unlearning in Section~\ref{sec:formulation}, we seek to remove a target fact $e$ and interrupt its reconstruction from knowledge retained by the pretrained model $\mathcal{M}_{\mathrm{pre}}$. As illustrated in Figure~\ref{fig:overview}, \textsc{DUMIC} consists of three stages: (1) Adaptive Tree Search combines entities explicitly generated in model responses with entities mapped from representations at the model's intermediate layers. These mapped entities are crucial for discovering supporting paths that response-only search misses, as shown in Section~\ref{sec:jlens_comparison};(2) Because discovered paths may be invalid, Supporting Path Pruning verifies and filters the candidates to form a supporting subgraph $\mathcal{G}_e$. (3) To interrupt the supporting paths while preserving as much other knowledge as possible, Minimum Cut Selection identifies a compact set of facts that intersects every path in $\mathcal{G}_e$ at minimum confidence-based cutting cost. 



\subsection{Discovering Supporting Paths via Adaptive Tree Search}
\label{sec:adaptive_search}

Given a target triple $e=(s^{*},r^{*},o^{*})$, our goal is to discover candidate supporting paths encoded in the pretrained model $\mathcal{M}_{\mathrm{pre}}$ by exploring relational connections from $s^{*}$ toward $o^{*}$ within a bounded path length. We organize exploration as a dynamically constructed search tree $\mathcal{T}=(\mathcal{V},\mathcal{E})$, where $\mathcal{V}$ contains search nodes and $\mathcal{E}$ records the expansion edges between them. Each node $n=(v,\pi)$ stores the current entity $v$ and an ordered sequence of relational triples $\pi$ connecting $s^{*}$ to $v$, with the root defined as $n_0=(s^{*},\varnothing)$. The same entity reached through different paths corresponds to distinct search nodes. Let $\mathcal{F}_d$ denote the frontier of nodes at depth $d$ awaiting expansion, each with a path of $d$ edges. Starting from $\mathcal{F}_0=\{n_0\}$, we perform breadth-first search by applying two stages to each node in the current frontier: candidate entity discovery and relational edge verification.

\noindent\textbf{Candidate Entity Discovery.}
For each node $n=(v,\pi)\in\mathcal{F}_d$, we query $\mathcal{M}_{\mathrm{pre}}$ for facts related to $v$ and extract the entities explicitly mentioned in its response. Since exhaustively identifying all related knowledge encoded in the model is infeasible, we seek to broaden search coverage by supplementing explicit responses with entity candidates decoded from intermediate representations. Specifically, we use the Jacobian lens (J-Lens)~\citep{gurnee2026verbalizable} to identify latent entities encoded in the model's intermediate representations but absent from its explicit response, providing additional candidates for extending $\pi$. However, decoding intermediate representations yields numerous candidate tokens, including noisy or uninterpretable fragments. To efficiently select relevant candidates, we first apply a probability threshold to reduce this noise, then rank the remaining tokens by their readout probabilities to prioritize candidates likely to be associated with the current entity $v$. We further resolve the selected tokens into entity names for subsequent relational edge verification.

\noindent\textbf{Relational Edge Verification.}
An entity's appearance in the response or intermediate representations does not establish its relational connection to $v$. For each candidate entity $w$, we therefore query $\mathcal{M}_{\mathrm{pre}}$ to determine whether a connecting relation exists and, if so, identify that relation. When a triple $(v,r,w)$ is returned, we append it to $\pi$ to obtain an extended path $\pi'$ and add a new search node $n'=(w,\pi')$ connected to $n$. Candidates for which no relation is obtained do not extend the search tree. To reduce search overhead, we simplify expansion at the final permitted hop. Since the next entity must be $o^{*}$ to complete a path within the remaining depth budget, we skip candidate entity discovery and directly verify the connection between each frontier entity $v$ and $o^{*}$. If a triple $(v,r,o^{*})$ is returned, we append it to the current path. This avoids further entity generation, intermediate-representation decoding, and relation queries for alternative candidates.

Whenever an extended path $\pi'$ reaches $o^{*}$ and satisfies the minimum path length, we record it as a completed candidate path. Otherwise, if the path remains eligible for expansion, we add the corresponding node $n'$ to the next frontier $\mathcal{F}_{d+1}$. Once all nodes in $\mathcal{F}_d$ have been processed, we repeat the expansion procedure for the next frontier. The search terminates when the maximum depth is reached or the frontier becomes empty. All completed candidate paths are then passed to the pruning stage to assess whether their constituent relations jointly support the target fact and to remove invalid or redundant paths.

\subsection{Pruning Supporting Paths for Subgraph Refinement}
\label{sec:path_pruning}

Connectivity between $s^*$ and $o^*$ does not ensure that the relations along a path support the target fact $e$. We therefore use LLM to assess whether the constituent triples jointly support the inference of $e$, restricting its judgment to the supplied facts. We discard unsupported paths and construct a supporting subgraph by taking the union of triples across the retained paths. We then determine which facts in this subgraph are retained by $\mathcal{M}_{\mathrm{pre}}$. To obtain reliable confidence estimates for this filtering step, we calibrate the model's output probabilities to better align its confidence with empirical accuracy. Each unique triple is converted into a multiple choice question with options A, B, C, D, and Unknown. A fact passes the filter only when the model selects the correct answer and the associated calibrated confidence satisfies the criterion.

\subsection{Interrupting Supporting Paths via Confidence-Aware Minimum Cut}
\label{sec:minimum_cut}

Effective deep unlearning requires not only preventing the recovery of target knowledge, but also preserving as much unrelated knowledge as possible. Directly unlearning every fact in $\mathcal{G}_e$ may reliably interrupt its supporting paths, but can cause unnecessary utility loss. This motivates selecting a compact subset of supporting facts that interrupts all recovery paths while minimizing the expected knowledge loss. We formulate supporting-fact selection as a confidence-aware minimum cut on $\mathcal{G}_e$, which excludes the target fact $e$. Removing a set of facts $C$ interrupts a path if at least one fact on that path belongs to $C$; $C$ constitutes a cut if its removal interrupts every path from $s^*$ to $o^*$ in $\mathcal{G}_e$. Let $\mathcal{K}_e$ denote the collection of such cuts. To account for the utility cost of removing supporting knowledge, for each supporting triple $(s,r,o)$, we query $\mathcal{M}_{\mathrm{pre}}$ about the triple and obtain its calibrated confidence $u_{\mathrm{pre}}(s,r,o)$ from the model's response. The corresponding fact is annotated as $t=(s,r,o,u)$, where $u=u_{\mathrm{pre}}(s,r,o)$, and assigned a cutting weight $c(t)=u$. This weighting uses confidence as a proxy for removal cost, favoring the preservation of more confidently held facts. Among all feasible cuts, we select the one with minimum total weight,
$C_e^* \in \arg\min_{C \in \mathcal{K}_e} \sum_{t \in C} c(t)$,
and combine its facts with $e$ to form the final forget set. The resulting cut interrupts all paths from $s^*$ to $o^*$ within the extracted subgraph at minimum total cutting cost.

\textbf{Discussion.}
The effectiveness of \textsc{DUMIC} depends on the coverage of its discovered supporting paths. A single search may miss valid paths because explicit responses reveal only part of the model's knowledge. Repeated searches with varied responses can uncover additional paths, but incur further inference and verification costs. To broaden coverage within a limited search budget, \textsc{DUMIC} augments explicitly generated entities with candidates mapped from representations at the model's intermediate layers~\citep{gurnee2026verbalizable} during a single search run. The resulting improvement in supporting-path discovery and target coverage is shown in Section~\ref{sec:jlens_comparison}. To reduce redundant computation, we reuse entity expansions across targets and cache direct tail checks for each combination of a target and an intermediate entity. At the final hop, we directly verify connections to the target tail entity, avoiding candidate generation, latent decoding, and relation checks for alternative entities across the final frontier. This eliminates an additional branching factor at the last expansion step, where the frontier can already be large. Concurrent model queries and batched latent decoding further improve throughput. Together, adaptive tree search takes only 80.2 seconds per target on average. Further implementation details and runtime measurements are provided in Appendix~\ref{app:path_search_efficiency}.

%% file: sections/5-evaluation.tex
\section{Evaluation}
\label{sec:search_evaluation}

\begin{figure}[t]
    \centering
    \vspace{-2mm}
    \includegraphics[width=\linewidth]{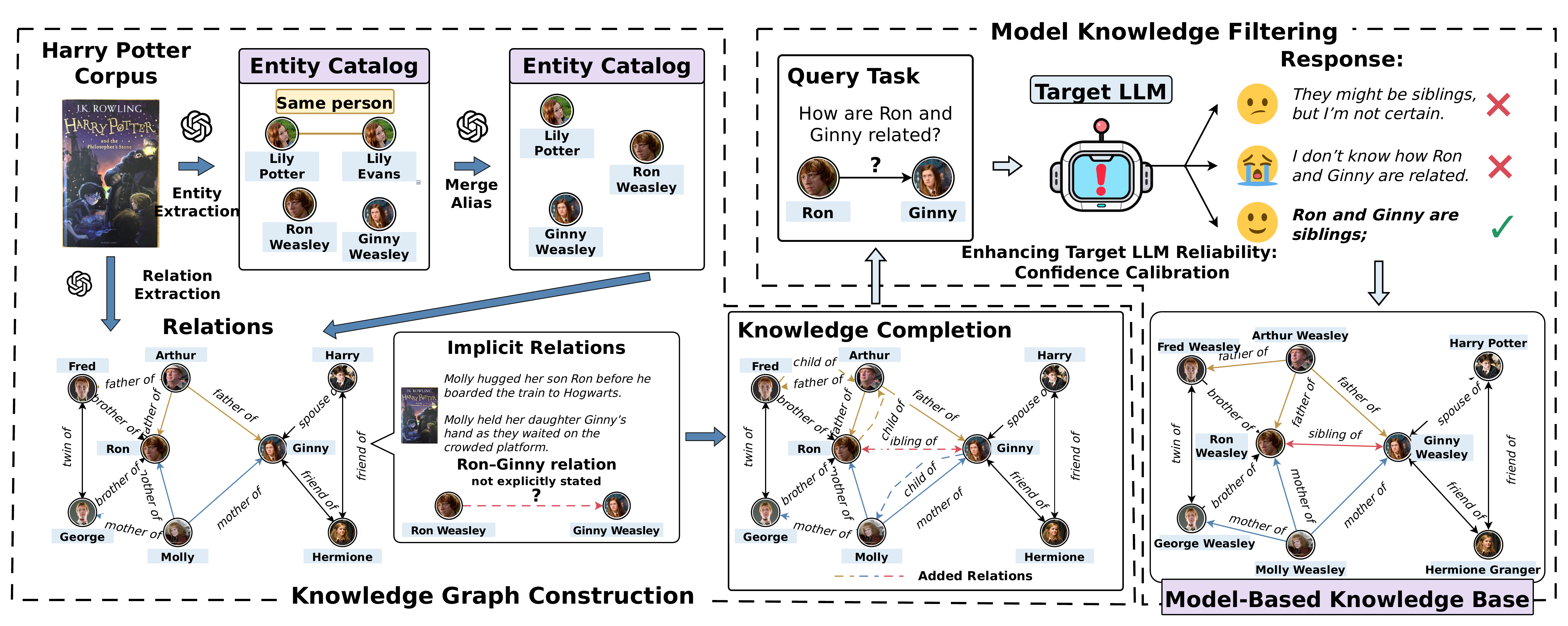}
    \vspace{-6mm}
    \caption{Construction of the model-based knowledge graph. We extract entities and relations from the Harry Potter corpus, merge aliases, and complete relations supported by the extracted facts. We then retain facts recognized by the target LLM according to its calibrated confidence.}
    \label{fig:kg_construction}
\end{figure}

To rigorously evaluate knowledge discovery and deep unlearning, we require a reference that approximates the model's internal knowledge structure, including the facts it retains and the inferential dependencies among them. Such a reference allows us to assess the validity and coverage of discovered supporting knowledge and identify target facts and their supporting structures for deep unlearning. We use the Harry Potter dataset~\citep{chen-etal-2023-large} as the starting point for constructing this reference. LLMs may have encountered the novels or related material during pretraining, making it plausible that they retain some of these facts. Relations among characters also provide rich inferential structures for examining whether target knowledge can be reconstructed from supporting facts. As illustrated in Figure~\ref{fig:kg_construction}, we construct and complete a reference knowledge graph from the corpus, then filter its facts using the pretrained model's responses and calibrated confidence.

\noindent\textbf{Constructing the Reference Knowledge Graph.}
To reduce omissions, knowledge extraction proceeds in two stages: entity extraction and relation extraction. The corpus contains numerous aliases, such as Lily Evans and Lily Potter for the same person. We consolidate canonical names and aliases into a unified entity catalog, preventing different mentions from creating separate nodes across chapters. We then extract relations from each chapter using its text and the entity catalog. To both identify missing relations and entities, we provide a second model with the same inputs and the initial extraction results for review.

We combine both outputs with their textual evidence and group relations by subject and object. Equivalent or overlapping descriptions, such as \textsc{SiblingOf} and \textsc{SisterOf} for the same entity pair, are consolidated by retaining the more specific relation supported by the evidence. To recover relations left implicit in the text, we examine each entity's neighborhood and supplement missing inverse relations and relations implied between neighboring entities. We record the supporting facts for each addition. These steps produce the reference KG $\mathcal{G}_{\mathrm{ref}}$.

\noindent\textbf{Filtering Facts According to Model Knowledge.}
We express each candidate triple as a multiple choice question with options A, B, C, D, and Unknown. We query the pretrained model $\mathcal{M}_{\mathrm{pre}}$ and obtain confidence $u$ from its calibrated answer probabilities. A fact passes the knowledge filter only when the model selects the correct answer and satisfies the confidence criterion. We apply the same probing procedure to target triples and candidate supporting facts, and use this procedure consistently during reference construction and evaluation after unlearning.

For each pretrained model, we first identify candidate target triples that pass the knowledge filter. Given a candidate $e=(s^*,r^*,o^*)$, we extract candidate paths connecting $s^*$ and $o^*$ in $\mathcal{G}_{\mathrm{ref}}$, excluding the target fact itself from the supporting evidence. For each path, we provide the target triple and the constituent facts to the judge, which determines whether these facts jointly support the inference of $e$. The judgment is restricted to the supplied facts, and paths that fail this assessment are removed. We then apply the knowledge filter to the constituent facts of each remaining path. A path survives only when all its facts pass this filter.






%% file: sections/6-experiments.tex

\section{Experiments}
\label{sec:experiments}
\subsection{Experimental Settings}
\label{sec:experimental_settings}

\paragraph{Experimental Setup}
\label{sec:experimental_setup}

We conduct experiments on Qwen3.5-27B~\citep{qwen2026qwen35} and Gemma4-31B-Base~\citep{gemmateam2026gemma4}, with a training budget of five epochs. We consider three unlearning settings: \textit{Superficial Unlearning}, which removes only the target facts; \textit{Deep Unlearning}, which removes the target facts and all facts in their discovered supporting subgraphs; and \textit{DUMIC Unlearning}, which removes the target facts and supporting facts selected by the confidence-aware minimum cut. We set the learning rate to $5\times10^{-6}$ for full parameter fine-tuning and $1\times10^{-5}$ for LoRA. For epoch selection, we select the last epoch whose accuracy on the utility set used for epoch selection exceeds 0.8; if no epoch satisfies this criterion, we select the epoch with the highest accuracy. We use GPT-5~\citep{openai2025gpt5} and Gemini 3.1 Pro~\citep{deepmind2026gemini31pro} to construct the reference knowledge graph. Gemini 3.1 Pro is additionally used for supporting-path search and evaluation dataset construction. We use the Jacobian lens (J-lens)~\citep{gurnee2026verbalizable} for latent entity discovery during search. Detailed implementation settings are provided in Appendix~\ref{app:experimental_details}.

\paragraph{Unlearning Methods}
For each target LLM, we evaluate five representative unlearning approaches. Gradient Ascent (GA)~\cite{jang2023knowledge} maximizes the prediction loss on the forget set to reduce the likelihood of the original responses. Negative Preference Optimization (NPO)~\cite{zhang2024negative} treats responses in the forget set as negative examples and reduces their likelihood relative to a frozen pretrained reference model. $\mathrm{GA}_{\mathrm{GDR}}$~\cite{yao2024machine} and $\mathrm{NPO}_{\mathrm{GDR}}$~\cite{zhang2024negative} combine their respective forgetting objectives with gradient descent on the prediction loss of the retain set to preserve model utility. Finally, Simple Negative Preference Optimization (SimNPO)~\cite{fan2024simplicity} uses a negative preference objective based on response log probabilities normalized by length, without requiring a reference model.

\paragraph{Evaluation Metrics.}
We evaluate deep unlearning with the Unlearning Effectiveness Score (UES). For a supporting fact $t$, let $C_m(t)$ be its calibrated confidence under model $m$; a path $P$ has confidence $C_m(P)=\min_{t\in P}C_m(t)$. Let $I(P)$ indicate whether any fact on $P$ receives zero confidence or an incorrect answer. We define $s(P)=I(P)+(1-I(P))[1-C_{\mathrm{unl}}(P)/C_{\mathrm{pre}}(P)]$ and $\mathrm{UES}=\frac{1}{N}\sum_{i=1}^{N}\frac{1}{|\mathcal{P}_i|}\sum_{P\in\mathcal{P}_i}s(P)$, where $N$ is the number of targets and $\mathcal{P}_i$ is the set of evaluated supporting paths for target $i$. For local utility, $\mathcal{N}_i$ contains 30 facts sampled within three hops of the target's subject or object, excluding the target and its supporting subgraph~\citep{wei2025forget}. We define $\mathrm{Loc}=\frac{1}{N}\sum_{i=1}^{N}\frac{1}{|\mathcal{N}_i|}\sum_{t\in\mathcal{N}_i}\mathbb{I}[\widehat{y}_{\mathrm{pre}}(t)=\widehat{y}_{\mathrm{unl}}(t)]$, where $\widehat{y}_m(t)$ is model $m$'s predicted answer. Gen and Rea assess general knowledge on MMLU~\citep{hendrycks2021measuring} and reasoning on BBH~\citep{suzgun2023challenging}, respectively. See Appendix~\ref{app:metric_details} for details.

\paragraph{Datasets}
\label{sec:datasets}

For each target model, we select the 100 recognized triples with the most verified supporting paths in the reference knowledge graph as unlearning targets. We then independently discover a model-specific supporting subgraph for each target using the search procedure in Section~\ref{sec:methodology}. We consider three forget sets: \textit{Unlearn} contains only the targets; \textit{DeepUnlearn} contains the targets and all facts in their searched supporting subgraphs; and \textit{DUMIC} contains the targets and the supporting facts selected by the minimum cut. We convert reference supporting facts and three-hop neighboring facts into A/B/C/D/Unknown multiple-choice questions with balanced gold-answer positions. Further details are provided in Appendix~\ref{app:dataset_construction}.

%% file: sections/7-results.tex
\subsection{Main Results}
\label{sec:main_results}

\begin{table*}[t]
\centering

\definecolor{unlearnblue}{RGB}{220,239,250}
\definecolor{utilityorange}{RGB}{255,231,216}
\definecolor{uesgreen}{RGB}{0,128,64}

\providecommand{\metrichead}[2]{%
    \begingroup
    \setlength{\fboxsep}{1.5pt}%
    \colorbox{#1}{\strut #2}%
    \endgroup
}

\providecommand{\ueschange}[4]{%
    #1\,{\scriptsize\textcolor{#4}{(#2\%$#3$)}}%
}

\caption{
Comparison of different unlearning settings on Qwen3.5-27B and
Gemma4-31B-Base. Percentages indicate relative UES improvements
over superficial unlearning.
}
\label{tab:unlearn_utility}

\setlength{\tabcolsep}{3pt}
\renewcommand{\arraystretch}{1.06}

\resizebox{\textwidth}{!}{%
\begin{tabular}{ll c ccc c ccc}
\toprule

\multirow{3}{*}{\textbf{Method}}
& \multirow{3}{*}{\textbf{Setting}}
& \multicolumn{4}{c}{\textbf{Qwen3.5-27B}}
& \multicolumn{4}{c}{\textbf{Gemma4-31B-Base}} \\

\cmidrule(lr){3-6}
\cmidrule(lr){7-10}

& & \metrichead{unlearnblue}{\textbf{Unlearning Effectiveness}}
& \multicolumn{3}{c}{
    \metrichead{utilityorange}{\textbf{Utility Retention}}
}
& \metrichead{unlearnblue}{\textbf{Unlearning Effectiveness}}
& \multicolumn{3}{c}{
    \metrichead{utilityorange}{\textbf{Utility Retention}}
} \\

\cmidrule(lr){3-3}
\cmidrule(lr){4-6}
\cmidrule(lr){7-7}
\cmidrule(lr){8-10}

& & \metrichead{unlearnblue}{\textbf{UES} $\uparrow$}
& \metrichead{utilityorange}{\textbf{Loc} $\uparrow$}
& \metrichead{utilityorange}{\textbf{Gen} $\uparrow$}
& \metrichead{utilityorange}{\textbf{Rea} $\uparrow$}
& \metrichead{unlearnblue}{\textbf{UES} $\uparrow$}
& \metrichead{utilityorange}{\textbf{Loc} $\uparrow$}
& \metrichead{utilityorange}{\textbf{Gen} $\uparrow$}
& \metrichead{utilityorange}{\textbf{Rea} $\uparrow$} \\

\midrule

\multicolumn{2}{l}{No Unlearn}
& 0.000 & 1.000 & 0.818 & 0.900
& 0.000 & 1.000 & 0.760 & 0.814 \\

\midrule

\multirow{3}{*}{GA}
& Superficial Unlearn
& 0.799 & 0.885 & 0.816 & 0.894
& 0.552 & 0.888 & 0.748 & 0.740 \\

& DeepUnlearn
& 0.963 & 0.707 & 0.814 & 0.874
& 1.000 & 0.000 & 0.738 & 0.000 \\

& DUMIC
& \ueschange{0.926}{15.89}{\uparrow}{uesgreen}
& 0.745 & 0.812 & 0.886
& \ueschange{0.690}{25.00}{\uparrow}{uesgreen}
& 0.905 & 0.756 & 0.698 \\

\midrule

\multirow{3}{*}{NPO}
& Superficial Unlearn
& 0.794 & 0.889 & 0.812 & 0.896
& 0.568 & 0.897 & 0.748 & 0.746 \\

& DeepUnlearn
& 0.965 & 0.647 & 0.814 & 0.878
& 1.000 & 0.000 & 0.732 & 0.000 \\

& DUMIC
& \ueschange{0.926}{16.62}{\uparrow}{uesgreen}
& 0.777 & 0.808 & 0.882
& \ueschange{0.700}{23.24}{\uparrow}{uesgreen}
& 0.900 & 0.742 & 0.712 \\

\midrule

\multirow{3}{*}{SimNPO}
& Superficial Unlearn
& 0.806 & 0.854 & 0.820 & 0.894
& 0.419 & 0.898 & 0.746 & 0.788 \\

& DeepUnlearn
& 0.946 & 0.652 & 0.806 & 0.886
& 0.809 & 0.790 & 0.730 & 0.706 \\

& DUMIC
& \ueschange{0.905}{12.28}{\uparrow}{uesgreen}
& 0.780 & 0.816 & 0.904
& \ueschange{0.687}{63.96}{\uparrow}{uesgreen}
& 0.905 & 0.746 & 0.738 \\

\midrule

\multirow{3}{*}{$\mathrm{GA}_{\mathrm{GDR}}$}
& Superficial Unlearn
& 0.839 & 0.873 & 0.812 & 0.896
& 0.107 & 0.976 & 0.758 & 0.744 \\

& DeepUnlearn
& 0.937 & 0.742 & 0.808 & 0.876
& 0.208 & 0.940 & 0.754 & 0.106 \\

& DUMIC
& \ueschange{0.936}{11.56}{\uparrow}{uesgreen}
& 0.781 & 0.818 & 0.880
& \ueschange{0.491}{358.88}{\uparrow}{uesgreen}
& 0.948 & 0.742 & 0.590 \\

\midrule

\multirow{3}{*}{$\mathrm{NPO}_{\mathrm{GDR}}$}
& Superficial Unlearn
& 0.555 & 0.948 & 0.816 & 0.880
& 0.387 & 0.926 & 0.762 & 0.778 \\

& DeepUnlearn
& 0.947 & 0.613 & 0.820 & 0.882
& 0.999 & 0.196 & 0.748 & 0.662 \\

& DUMIC
& \ueschange{0.883}{59.10}{\uparrow}{uesgreen}
& 0.891 & 0.816 & 0.886
& \ueschange{0.583}{50.65}{\uparrow}{uesgreen}
& 0.844 & 0.758 & 0.744 \\

\bottomrule
\end{tabular}%
}
\end{table*}

\paragraph{Deep Unlearning Strengthens Forgetting}
\label{sec:main_unlearning_comparison}

Table~\ref{tab:unlearn_utility} shows that DeepUnlearn consistently improves UES over superficial unlearning across both models. For example, on Gemma-4-31B-Base, $\mathrm{NPO}_{\mathrm{GDR}}$ improves from $0.387$ to $0.999$. On Qwen3.5-27B, DeepUnlearn achieves UES above $0.93$ across all five methods, indicating that explicitly targeting supporting facts weakens recovery paths more effectively than unlearning the targets alone. However, removing the entire supporting subgraph can severely damage utility: on Gemma-4-31B-Base, $\mathrm{NPO}_{\mathrm{GDR}}$ reduces Loc from $0.926$ to $0.196$, while GA and NPO reduce both Loc and Rea to zero. These results show that removing supporting knowledge strengthens forgetting but may incur substantial utility loss.

\paragraph{Utility Preservation through Minimum-Cut Selection}
\label{sec:mincut_comparison}

Minimum-cut selection mitigates this utility loss while maintaining
higher UES than superficial unlearning across all five methods on
both models. On Gemma-4-31B-Base, it restores GA's Loc and Rea
from zero under DeepUnlearn to $0.905$ and $0.698$, respectively.
For $\mathrm{NPO}_{\mathrm{GDR}}$, it improves Loc from $0.196$
to $0.844$, while achieving a UES of $0.583$, compared with
$0.387$ under superficial unlearning. Methods with lower superficial
UES often obtain larger gains, potentially because explicitly
targeting supporting facts addresses recovery paths left largely
intact by target-only unlearning. Lower baselines also leave more
room for improvement and amplify relative percentage gains.

\paragraph{Effectiveness of Supporting Path Discovery}
\label{sec:supporting_path_quantity}

\begin{wrapfigure}[19]{r}{0.30\textwidth}
    \centering
    \vspace{-2.0\baselineskip}    \includegraphics[width=\linewidth]{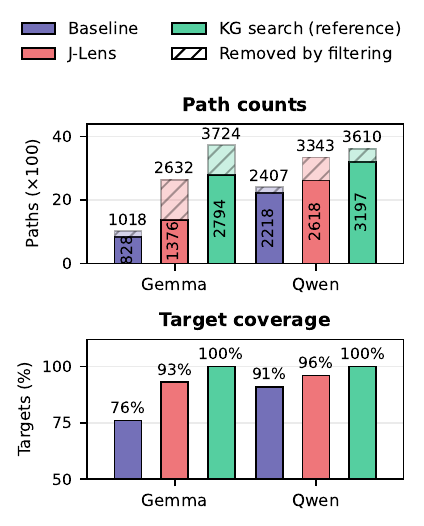}
\caption{Supporting path discovery. Top: path counts before and after filtering, with hatched segments indicating removed paths. Bottom: target coverage after filtering.}
    \label{fig:supporting_path_discovery}
\end{wrapfigure}

As shown in Figure~\ref{fig:supporting_path_discovery}, J-Lens discovers 2,632 candidate paths on Gemma and 3,343 on Qwen, of which 1,376 and 2,618 remain after filtering, respectively. These retained paths cover 93\% of targets on Gemma and 96\% on Qwen, showing that the procedure identifies supporting paths for most targets. For comparison, searching the reference KG yields 2,794 retained paths on Gemma and 3,197 on Qwen, covering all targets on both models. Although model-guided search does not exhaust the supporting paths in the reference KG, J-Lens discovers substantial supporting evidence without using the reference KG to guide its search.
\paragraph{Contribution of the Jacobian Lens}
\label{sec:jlens_comparison}

We next compare J-Lens with the baseline to assess the contribution of
the Jacobian lens. Before filtering, J-Lens increases the number of
discovered paths from 1,018 to 2,632 on Gemma and from 2,407 to 3,343 on
Qwen, corresponding to 2.59$\times$ and 1.39$\times$ the baseline counts,
respectively. The gains persist after filtering: the number of retained
paths increases from 828 to 1,376 on Gemma (+66.2\%) and from 2,218 to
2,618 on Qwen (+18.0\%). Target coverage also improves, rising from
76\% to 93\% on Gemma and from 91\% to 96\% on Qwen.  These improvements show that the benefits of J-Lens extend to both the number of retained supporting
paths and the fraction of targets covered. 

\par
\par\wrapfill

\subsection{Reliability of Model Confidence Estimates}
\label{sec:calibration}

\begin{figure}[!t]
    \centering
    \label{app:calibration_diagrams}\includegraphics[width=\linewidth]{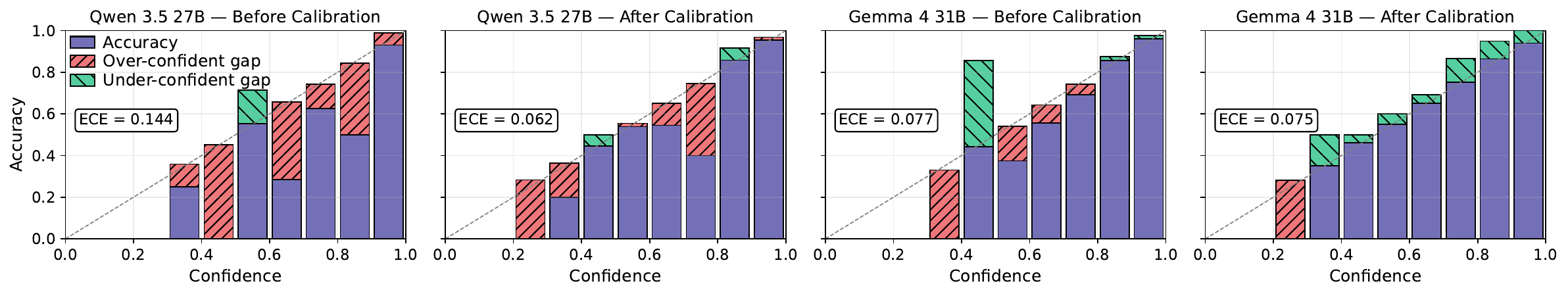}
    \caption{Reliability diagrams before and after temperature scaling.}
    \label{fig:calibration}
\end{figure}

To assess the reliability of model confidence for knowledge evaluation, we apply temperature scaling to the answer token probabilities. Using a calibration set containing 200 positive and 200 negative triples, we fit temperatures of 1.6 for Qwen3.5-27B and 1.3 for Gemma4-31B-Base. We evaluate calibration on a separate set of 100 positive and 100 negative triples using expected calibration error and reliability diagrams. As illustrated in Figure~\ref{fig:calibration}, calibration reduces the expected calibration error of Qwen3.5-27B from 0.144 to 0.062, indicating improved alignment between predicted confidence and observed accuracy. Gemma4-31B-Base exhibits a lower initial calibration error, with a modest reduction from 0.077 to 0.075. We subsequently use the calibrated probabilities to compute entropy for knowledge filtering and supporting subgraph evaluation.

\subsection{Ablation Studies}
\label{sec:ablation}

\paragraph{Effectiveness of Deep Unlearning under Low-Rank Adaptation}
\label{sec:ablation_finetuning_lora}

Table~\ref{tab:lora_unlearn_utility} shows that DUMIC improves UES across all five methods under LoRA.
The gains are particularly large for GA and $\mathrm{GA}_{\mathrm{GDR}}$:
GA improves from $0.109$ to $0.789$, while
$\mathrm{GA}_{\mathrm{GDR}}$ improves from $-0.643$ to $0.548$.
$\mathrm{GA}_{\mathrm{GDR}}$ achieves these gains with modest changes
in Loc, Gen, and Rea, whereas GA incurs larger losses in local utility
and reasoning performance. These results demonstrate that our framework also applies to LoRA, improving unlearning effectiveness under parameter-efficient adaptation.

\begin{table*}[t]
\centering
\definecolor{uesgreen}{RGB}{0,128,64}
\definecolor{uesred}{RGB}{190,35,45}

\providecommand{\ueschange}[4]{%
    #1\,{\scriptsize\textcolor{#4}{(#2\%$#3$)}}%
}

\caption{
Comparison of unlearning methods using LoRA on Qwen3.5-27B under
superficial unlearning and deep unlearning with minimum cut.
}
\label{tab:lora_unlearn_utility}
\setlength{\tabcolsep}{2.2pt}
\renewcommand{\arraystretch}{1.03}

\scriptsize
\resizebox{0.96\textwidth}{!}{%
\begin{tabular}{l cccc cccc}
\toprule

\multirow{3}{*}{Method}
& \multicolumn{4}{c}{Superficial Unlearning}
& \multicolumn{4}{c}{DUMIC} \\

\cmidrule(lr){2-5}
\cmidrule(lr){6-9}

& Unlearning Effectiveness
& \multicolumn{3}{c}{Utility Retention}
& Unlearning Effectiveness
& \multicolumn{3}{c}{Utility Retention} \\

\cmidrule(lr){2-2}
\cmidrule(lr){3-5}
\cmidrule(lr){6-6}
\cmidrule(lr){7-9}

& UES $\uparrow$
& Loc $\uparrow$
& Gen $\uparrow$
& Rea $\uparrow$
& UES $\uparrow$
& Loc $\uparrow$
& Gen $\uparrow$
& Rea $\uparrow$ \\

\midrule

Base Model
& 0.000 & 1.000 & 0.818 & 0.900
& 0.000 & 1.000 & 0.818 & 0.900 \\

\midrule

GA
& 0.109 & 0.923 & 0.812 & 0.834
& \ueschange{0.789}{623.85}{\uparrow}{uesgreen}
& 0.730 & 0.802 & 0.622 \\

NPO
& -0.168 & 0.935 & 0.812 & 0.846
& \ueschange{0.337}{300.60}{\uparrow}{uesgreen}
& 0.914 & 0.812 & 0.776 \\

SimNPO
& 0.083 & 0.924 & 0.818 & 0.894
& \ueschange{0.141}{69.88}{\uparrow}{uesgreen}
& 0.944 & 0.812 & 0.886 \\

$\mathrm{GA}_{\mathrm{GDR}}$
& -0.643 & 0.936 & 0.808 & 0.706
& \ueschange{0.548}{185.23}{\uparrow}{uesgreen}
& 0.911 & 0.806 & 0.686 \\

$\mathrm{NPO}_{\mathrm{GDR}}$
& -0.377 & 0.962 & 0.816 & 0.864
& \ueschange{-0.051}{86.47}{\uparrow}{uesgreen}
& 0.957 & 0.816 & 0.882 \\

\bottomrule
\end{tabular}%
}
\end{table*}

\begin{figure}[t]
    \centering
    \includegraphics[
        width=1.0\linewidth]{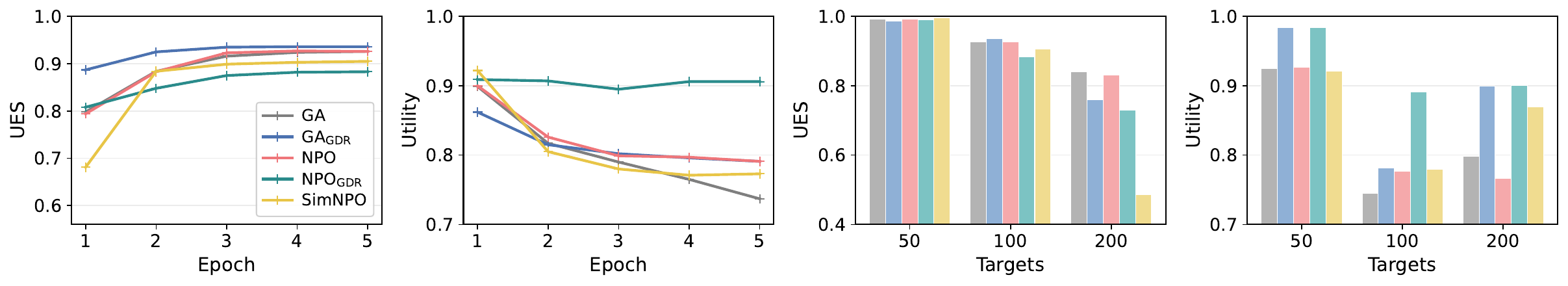}
\caption{Effects of unlearning epochs (left) and target count (right) on UES and utility for Qwen3.5-27B with full-parameter fine-tuning and DUMIC.}
    \label{fig:analysis}
\end{figure}

\paragraph{Impact of Unlearning Iterations}
\label{sec:epoch_analysis}

We analyze five unlearning epochs on Qwen3.5-27B under DUMIC.
As shown in Figure~\ref{fig:analysis}, UES improves
mainly during the first two to three epochs and then
largely stabilizes. SimNPO exhibits the largest early gain,
with UES increasing from approximately $0.68$ to $0.88$
between the first and second epochs.
Local utility generally declines with further training;
for example, GA's Loc decreases from approximately $0.92$
to $0.74$. These results indicate diminishing forgetting gains
in later epochs, often accompanied by additional utility loss.

\paragraph{Impact of Target Count}
\label{sec:target_count}

The right two panels of Figure~\ref{fig:analysis} compare unlearning the top 50, 100, and 200 targets, ranked by the number of verified supporting paths in the reference knowledge graph. Average UES decreases from 0.9913 to 0.9152 and 0.7294, respectively. This decline appears across all five methods: from 50 to 200 targets, GA falls from 0.9921 to 0.8412, while SimNPO falls from 0.9962 to 0.4852. Thus, forgetting becomes less effective as the target set expands. This decline may partly reflect lower supporting-path coverage for the additional targets, leaving more recovery paths outside the discovered subgraphs and therefore beyond the reach of minimum-cut selection.








%% file: sections/8-conclusion.tex

\section{Conclusion}

We introduced a deep unlearning framework that discovers model-specific supporting paths through adaptive tree search and latent entity extraction from intermediate representations. After verifying these paths and constructing a supporting subgraph, the framework uses a confidence-aware minimum cut to select facts for unlearning while limiting collateral knowledge loss. Experiments on two LLMs show improved path coverage and a better balance between unlearning effectiveness and utility preservation. These findings highlight the importance of addressing related knowledge that can reconstruct a target fact.

%% file: sections/9-ai-use-statement.tex


\section{Acknowledgement}
Rongzhe Wei, Pan Li, Hans Hao-Hsun Hsu are partially supported by the National Science Foundation (NSF) under awards PHY-2117997,
IIS-2239565, IIS-2428777, and CCF-2402816; the NAIRR Pilot projects 250459 and 250487; the Google Cloud Research Credit, 2026; the Nvidia Academic Award, 2026; and the Amazon Academic Award, 2026;  

%% file: sections/A-appendix-experimental-settings.tex

\section{Detailed Experimental Settings}
\label{app:experimental_details}

\subsection{Compute Configuration}
\label{app:compute_configurations}

All experiments were conducted on NVIDIA A100 80GB PCIe GPUs,
which were used for both training and inference.

\subsection{Search Configuration}
\label{app:search_configurations}

We use \texttt{vLLM serve}~\citep{kwon2023efficient} for efficient model inference during supporting-path search, with batch sizes of 512 for Qwen3.5-27B and 256 for Gemma4-31B-Base. We compare two search settings in Figure~\ref{fig:supporting_path_discovery}: the baseline uses only entities explicitly generated in model responses, whereas J-Lens additionally incorporates latent entities decoded from intermediate representations.

\subsection{J-Lens Configuration}
We apply J-Lens~\citep{gurnee2026verbalizable} to token positions within the first three object fields of each probe response, using only the last of the three inspected layers for candidate extraction. Within each field, we aggregate readout probabilities by taking the maximum across its token positions. To reduce noise from spurious token fragments, we retain up to three new latent entities per field with readout probabilities of at least $10^{-4}$. Each expansion admits at most 50 candidates, with explicitly generated entities taking priority over latent candidates. 

\subsection{Training Configuration}
\label{app:training_configurations}

All unlearning experiments were performed using four GPUs
with DeepSpeed ZeRO Stage~3~\citep{rajbhandari2020zero}.
The per-GPU batch size was set to 4. Selected training epochs for all combinations of models, methods, and unlearning settings are reported in Table~\ref{tab:epoch_selection}.

\begin{table}[htbp]
    \centering
    \caption{Selected training epochs under different unlearning
    settings using full-parameter fine-tuning (Full) and LoRA.
    Sup., Deep, and +Cut denote superficial unlearning, deep
    unlearning, and deep unlearning with minimum cut.}
    \label{tab:epoch_selection}
    \small
    \setlength{\tabcolsep}{6pt}
    \renewcommand{\arraystretch}{1.12}
    \begin{tabular}{l ccc ccc ccc}
        \toprule
        \multirow{3}{*}{\textbf{Method}}
        & \multicolumn{6}{c}{\textbf{Qwen3.5-27B}}
        & \multicolumn{3}{c}{\textbf{Gemma4-31B-Base}} \\
        \cmidrule(lr){2-7} \cmidrule(lr){8-10}
        & \multicolumn{3}{c}{Full} & \multicolumn{3}{c}{LoRA}
        & \multicolumn{3}{c}{Full} \\
        \cmidrule(lr){2-4} \cmidrule(lr){5-7} \cmidrule(lr){8-10}
        & Sup. & Deep & +Cut & Sup. & Deep & +Cut & Sup. & Deep & +Cut \\
        \midrule
        GA                            & 5 & 5 & 5 & 5 & 5 & 5 & 2 & 1 & 1 \\
        NPO                           & 5 & 5 & 5 & 5 & 5 & 5 & 2 & 1 & 1 \\
        SimNPO                        & 5 & 5 & 5 & 5 & 5 & 5 & 5 & 5 & 5 \\
        $\mathrm{GA}_{\mathrm{GDR}}$  & 5 & 5 & 5 & 5 & 1 & 5 & 1 & 1 & 1 \\
        $\mathrm{NPO}_{\mathrm{GDR}}$ & 5 & 5 & 5 & 5 & 5 & 5 & 5 & 5 & 5 \\
        \bottomrule
    \end{tabular}
\end{table}

\subsection{Detailed Definitions of Evaluation Metrics}
\label{app:metric_details}

Let $\mathcal{D}_{\mathrm{forget}}=\{e_i\}_{i=1}^{N}$ denote the set of
target triples. For each target $e_i$, let
$\mathcal{G}_i=(\mathcal{V}_i,\mathcal{E}_i)$ denote its supporting
subgraph, where $\mathcal{E}_i$ contains the unique facts in the union of
all verified supporting paths. We use $\mathcal{P}_i$ to denote the set
of supporting paths for $e_i$. The pretrained and unlearned models are
denoted by $\mathcal{M}_{\mathrm{pre}}$ and
$\mathcal{M}_{\mathrm{unl}}$, respectively.

\paragraph{Unlearning Effectiveness Score (UES).}
UES measures the reduction in confidence of supporting paths after
unlearning. For each supporting fact $t$, let $C_{\mathrm{pre}}(t)$
and $C_{\mathrm{unl}}(t)$ denote its calibrated confidence under the
pretrained and unlearned models, respectively. Since reconstructing
the target through a path requires all its constituent facts, we
approximate path confidence by the lowest confidence among these facts:
\begin{equation}
    C_m(P) = \min_{t \in P} C_m(t),
    \qquad m \in \{\mathrm{pre},\mathrm{unl}\}.
    \label{eq:path_confidence}
\end{equation}
We consider a path interrupted if, after unlearning, at least one
constituent fact has zero confidence or is answered incorrectly,
including an Unknown response. Such a path receives a score of $1$.
Otherwise, its score is the normalized reduction in path confidence:
\begin{equation}
    s(P) =
    \begin{cases}
        1,
        & \text{if } \exists\, t \in P:
          C_{\mathrm{unl}}(t)=0
          \text{ or } \hat{a}_{\mathrm{unl}}(t)\neq a^*(t), \\[4pt]
        \displaystyle
        \frac{C_{\mathrm{pre}}(P)-C_{\mathrm{unl}}(P)}
             {C_{\mathrm{pre}}(P)},
        & \text{otherwise},
    \end{cases}
    \label{eq:path_ues}
\end{equation}
where $\hat{a}_{\mathrm{unl}}(t)$ and $a^*(t)$ denote the unlearned
model's predicted answer and the correct answer for fact $t$,
respectively. We evaluate paths with positive pretrained confidence.
UES averages these scores over supporting paths and then over targets:
\begin{equation}
    \mathrm{UES}
    =
    \frac{1}{N}
    \sum_{i=1}^{N}
    \frac{1}{|\mathcal{P}_i|}
    \sum_{P \in \mathcal{P}_i} s(P).
    \label{eq:ues}
\end{equation}
A higher UES indicates more effective interruption or weakening of
supporting paths. A value of $1$ indicates that every evaluated path
is interrupted under the above criterion, while a negative value
indicates that confidence increases in the remaining paths outweigh
the contributions from interrupted or weakened paths.

\paragraph{Local Consistency (Loc).}
Loc evaluates whether unlearning preserves neighboring factual knowledge
that is not involved in reconstructing the target. For each target
$e_i=(s_i,r_i,o_i)$, we sample 30 facts from the three-hop neighborhoods
of $s_i$ and $o_i$ in the reference knowledge graph. We exclude the target
itself and all facts in its supporting subgraph. Let $\mathcal{N}_i$
denote the resulting local utility set, and let
$\widehat{y}_{\mathcal{M}}(t)$ denote the model's prediction among
$\{\textsc{A},\textsc{B},\textsc{C},\textsc{D},\textsc{Unknown}\}$.
Loc is defined as
\begin{equation}
    \mathrm{Loc}
    =
    \frac{1}{N}
    \sum_{i=1}^{N}
    \frac{1}{|\mathcal{N}_i|}
    \sum_{t\in\mathcal{N}_i}
    \mathbb{I}
    \left[
        \widehat{y}_{\mathrm{pre}}(t)
        =
        \widehat{y}_{\mathrm{unl}}(t)
    \right].
    \label{eq:loc}
\end{equation}
A higher Loc indicates better preservation of local factual
knowledge around the unlearning targets.

\paragraph{General Knowledge and Reasoning Ability.}
General Knowledge (Gen) is measured on
MMLU~\citep{hendrycks2021measuring} using 5-shot perplexity-based answer
ranking. Reasoning Ability (Rea) is measured on the 27 tasks in
BBH~\citep{suzgun2023challenging} using 3-shot chain-of-thought prompting
and exact-match accuracy. Higher Gen and Rea indicate better preservation
of general knowledge and reasoning capabilities, respectively.

%% file: sections/B-appendix-dataset-construction.tex

\section{Dataset Construction Details}
\label{app:dataset_construction}

\paragraph{Unlearning Targets and Supporting Subgraphs.}
We select the 100 target triples with the largest numbers of verified supporting paths in the reference knowledge graph constructed in Section~\ref{sec:search_evaluation}. We define 19 sentence templates and randomly sample one for each target triple, converting it into a natural-language sentence for unlearning. For each target $e_i$, the discovered paths and their facts form a model-specific searched subgraph $\widehat{\mathcal{G}}_i$, which is used to construct the forget sets and perform deep unlearning. The resulting forget-set sizes and retain sets are summarized in Table~\ref{tab:training_set_sizes}. We further retrieve the corresponding supporting paths from the reference knowledge graph and combine them into a reference subgraph $\mathcal{G}_i^{\mathrm{ref}}$. This fixed, method-independent subgraph is used to measure fact-level and path-level unlearning effectiveness.

\begin{table}[htbp]
\centering
\caption{
Numbers of forget and retain examples used under different unlearning
settings. Each retain set is matched in size to its corresponding forget set.
}
\label{tab:training_set_sizes}
\small
\setlength{\tabcolsep}{7pt}
\renewcommand{\arraystretch}{1.10}
\begin{tabular}{lcccccc}
\toprule
\multirow{2}{*}{\textbf{Model}}
& \multicolumn{2}{c}{\textbf{Unlearn}}
& \multicolumn{2}{c}{\textbf{DUMIC}}
& \multicolumn{2}{c}{\textbf{DeepUnlearn}} \\
\cmidrule(lr){2-3} \cmidrule(lr){4-5} \cmidrule(lr){6-7}
& Forget & Retain & Forget & Retain & Forget & Retain \\
\midrule
Qwen3.5-27B     & 100 & 100 & 236 & 236 & 616 & 616 \\
Gemma4-31B-Base & 100 & 100 & 232 & 232 & 639 & 639 \\
\bottomrule
\end{tabular}
\end{table}

\paragraph{Multiple-Choice Evaluation.}
We prompt an LLM to convert both the facts in the reference supporting
subgraphs and the local utility facts into four-option multiple-choice
questions. Each question contains one gold answer and three plausible
distractors, with gold positions evenly balanced across A--D. The
supporting-subgraph questions are used to compute UES. For
Loc, we sample 30 facts from the three-hop neighborhoods of each target's
head and tail entities, excluding the target and its supporting facts. We compare
model predictions on these fixed questions before and after unlearning.
For epoch selection, we randomly sample
100 questions from a separate utility set.

\paragraph{Confidence Calibration Dataset.}
We construct separate calibration sets for each target model. The
temperature-fitting set contains 200 positive and 200 negative examples,
while the held-out calibration evaluation set contains another 100
positive and 100 negative examples. All examples are presented using the
same A/B/C/D/Unknown format as the effectiveness and utility evaluations.
We fit a separate temperature for each model on the fitting set and report
calibration quality on the held-out set. The calibration examples are
separate from the unlearning targets and their effectiveness and utility
evaluation questions.

%% file: sections/C-appendix-kg-statistics.tex

\section{Knowledge Graph Statistics}
\label{app:kg_statistics}

The Harry Potter series provides a coherent fictional world with rich
character relationships and multi-hop factual dependencies. We use the complete English texts of all seven original Harry Potter
books as our corpus, following the fictional domain studied by the Harry
Potter Dialogue dataset, a bilingual collection of character-centered
dialogues~\citep{chen-etal-2023-large}.

\begin{figure}[htbp]
    \centering
    \includegraphics[width=\linewidth]{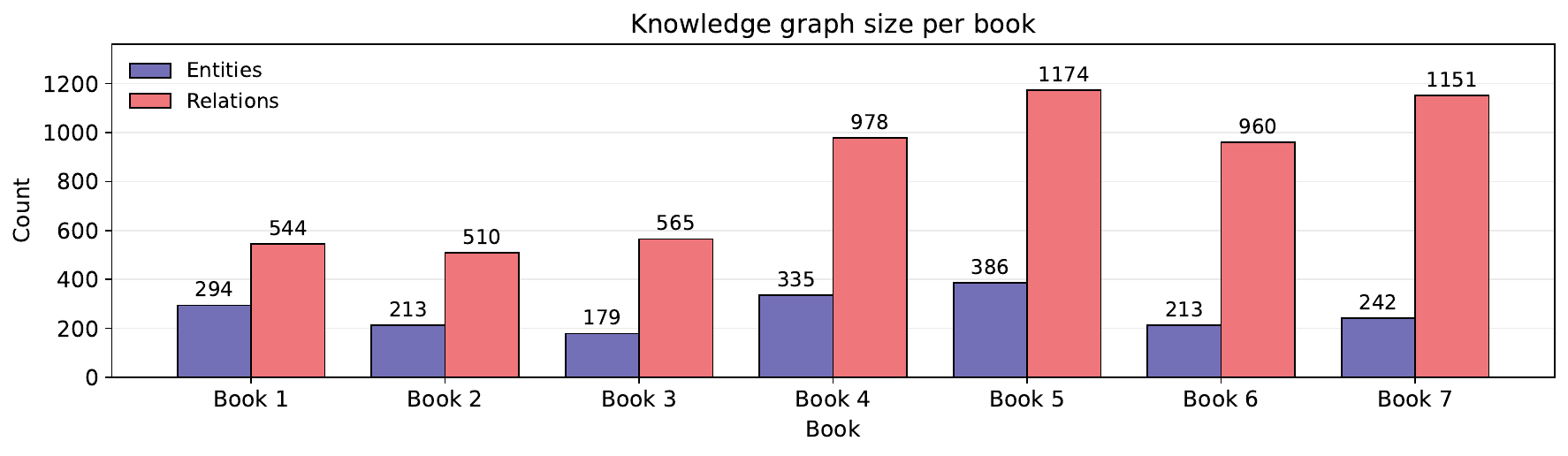}
    \caption{Numbers of entities and relation triples extracted
    from each of the seven Harry Potter books.}
    \label{fig:kg_num}
\end{figure}

Figure~\ref{fig:kg_num} summarizes the numbers of entities and
relation triples extracted from each of the seven Harry Potter
books. The per-book extractions contain between 179 and 386
entities and between 510 and 1,174 relation triples, with Book~5
yielding the largest counts in both categories. After consolidating
the extracted knowledge and removing duplicates, the resulting
reference knowledge graph contains 1,862 entities and 4,122
relation triples.

%% file: sections/D-appendix-search-efficiency.tex

\section{Efficiency of Supporting Path Discovery}
\label{app:path_search_efficiency}

We use vLLM~\citep{kwon2023efficient} for model inference during
supporting-path search.
Table~\ref{tab:path-search-timing} reports the runtime and path
statistics for 200 target facts. On Qwen3.5-27B, our framework
requires an average of 80.2 seconds per target, discovering
54.9 candidate paths and retaining 20.3 accepted paths per target.
On Gemma4-31B-Base, the corresponding runtime is 157.9 seconds,
with averages of 46.9 candidate paths and 15.9 accepted paths
per target. Thus, the average search time remains below three
minutes per target on both models at the evaluated scale.

\begin{table}[htbp]
    \centering
    \caption{Runtime and path statistics of supporting-path
    discovery over 200 target facts. Candidate and accepted
    path counts are totals across all targets.}
    \label{tab:path-search-timing}
    \small
    \setlength{\tabcolsep}{8pt}
    \renewcommand{\arraystretch}{1.10}
    \begin{tabular}{lrrrr}
        \toprule
        \textbf{Model}
        & \textbf{Total time}
        & \textbf{Avg.\ per target}
        & \textbf{Candidate paths}
        & \textbf{Accepted paths} \\
        \midrule
        Qwen3.5-27B
        & 4.45\,h & 80.2\,s & 10{,}971 & 4{,}067 \\
        Gemma4-31B-Base
        & 8.77\,h & 157.9\,s & 9{,}370 & 3{,}181 \\
        \bottomrule
    \end{tabular}
\end{table}



%% file: sections/E-appendix-entropy-mapping.tex

\section{Mapping Between Entropy Threshold and Correct-Answer Probability}
\label{app:entropy_probability}

In this section, we explain how the entropy-based filtering
criterion used for constructing supporting subgraphs relates
to the model's probability of selecting the correct answer.
We consider five options,
$\mathcal{A}=\{A,B,C,D,\mathrm{Unknown}\}$, and denote the
correct option by $a^*\in\{A,B,C,D\}$.
Let $\mathbf{q}$ be the model's probability distribution
normalized over these five options.
A prediction is considered to reflect retained knowledge
when (i) the correct option has the highest probability and
(ii) the entropy of $\mathbf{q}$ is below a threshold $u^*$.
We compute entropy in bits as
$H(\mathbf{q})=-\sum_{a\in\mathcal{A}} q_a\log_2 q_a$.

Table~\ref{tab:entropy_probability} reports the feasible
ranges of the correct-answer probability $p=q_{a^*}$ at
different fixed entropy values $H(\mathbf{q})=u^*$,
subject to the correct option being an argmax.
The upper bound occurs when the other four options share
the remaining probability equally:
\begin{equation}
    u^* =
    H\left(
        p,\frac{1-p}{4},\frac{1-p}{4},
        \frac{1-p}{4},\frac{1-p}{4}
    \right),
    \qquad p \geq \frac{1-p}{4}.
\end{equation}
For the range $0\leq u^*\leq 1$ bit considered here,
the lower bound occurs when one alternative receives
all the remaining probability and the other three
receive zero:
\begin{equation}
    u^* = H(p,1-p,0,0,0),
    \qquad p \geq 1-p.
\end{equation}
These intervals characterize the boundary $H(\mathbf{q})=u^*$.
Under the actual filtering condition $H(\mathbf{q})<u^*$,
the correct-answer probability exceeds the corresponding
lower bound and may reach $1$.

\begin{table}[t]
    \centering
    \caption{
        Feasible correct-answer probability ranges at fixed
        entropy values $H(\mathbf{q})=u^*$ for five options
        (A, B, C, D, and Unknown), assuming the correct option
        has the highest probability. Entropy is measured in bits.
    }
    \label{tab:entropy_probability}
    \small
    \setlength{\tabcolsep}{6pt}
    \renewcommand{\arraystretch}{1.10}
    \begin{tabular}{@{}cc@{\hspace{2.5em}}cc@{}}
        \toprule
        \textbf{Entropy} & \textbf{Correct Prob.\ Range}
        & \textbf{Entropy} & \textbf{Correct Prob.\ Range} \\
        \midrule
        0.10 & $[0.987, 0.990]$ & 0.60 & $[0.854, 0.913]$ \\
        0.15 & $[0.978, 0.984]$ & 0.65 & $[0.833, 0.904]$ \\
        0.20 & $[0.969, 0.978]$ & 0.70 & $[0.811, 0.894]$ \\
        0.25 & $[0.958, 0.971]$ & 0.75 & $[0.785, 0.884]$ \\
        0.30 & $[0.947, 0.963]$ & 0.80 & $[0.757, 0.874]$ \\
        0.35 & $[0.934, 0.956]$ & 0.85 & $[0.724, 0.863]$ \\
        0.40 & $[0.921, 0.948]$ & 0.90 & $[0.684, 0.852]$ \\
        0.45 & $[0.906, 0.940]$ & 0.95 & $[0.631, 0.841]$ \\
        0.50 & $[0.890, 0.931]$ & 1.00 & $[0.500, 0.829]$ \\
        0.55 & $[0.873, 0.922]$ &      &                   \\
        \bottomrule
    \end{tabular}
\end{table}

%% file: sections/F-appendix-additional-ablations.tex

\section{Additional Ablation Studies}
\label{app:additional_ablations}

\subsection{Controlling for Forget-Set Size}
\label{sec:ablation_forget_set_size}

To determine whether the gains of minimum-cut selection arise merely from using more training examples, we expand each superficial forget set to match the corresponding minimum-cut set of 236 examples and train both settings for five epochs. As shown in Table~\ref{tab:forget_set_size_ablation}, DUMIC achieves higher UES across all five methods, with absolute gains ranging from 0.033 to 0.167, although Loc decreases in every case. The largest UES gain occurs for $\mathrm{NPO}_{\mathrm{GDR}}$, whose UES increases from 0.716 to 0.883 while Loc decreases from 0.924 to 0.891. These results show that the forgetting gains cannot be explained by forget-set size alone: selecting supporting facts improves unlearning effectiveness under the same example budget, at the cost of some local utility.

\begin{table*}[t]
    \centering
    \caption{
        Superficial unlearning versus DUMIC on Qwen3.5-27B
        with 236 forget examples and five training epochs.
        Higher UES is bolded.
    }
    \label{tab:forget_set_size_ablation}
    \small
    \setlength{\tabcolsep}{6pt}
    \renewcommand{\arraystretch}{1.06}
    \begin{tabular}{@{}lcccc@{}}
        \toprule
        \multirow{2}{*}{\textbf{Method}}
        & \multicolumn{2}{c}{\metrichead{unlearnblue}{\textbf{UES} $\uparrow$}}
        & \multicolumn{2}{c}{\metrichead{utilityorange}{\textbf{Loc} $\uparrow$}} \\
        \cmidrule(lr){2-3}
        \cmidrule(lr){4-5}
        & Superficial & DUMIC
        & Superficial & DUMIC \\
        \midrule
        GA
        & 0.871 & \textbf{0.926}
        & 0.857 & 0.745 \\
        $\mathrm{GA}_{\mathrm{GDR}}$
        & 0.871 & \textbf{0.936}
        & 0.863 & 0.781 \\
        NPO
        & 0.893 & \textbf{0.926}
        & 0.870 & 0.777 \\
        $\mathrm{NPO}_{\mathrm{GDR}}$
        & 0.716 & \textbf{0.883}
        & 0.924 & 0.891 \\
        SimNPO
        & 0.869 & \textbf{0.905}
        & 0.817 & 0.780 \\
        \bottomrule
    \end{tabular}
\end{table*}